\pdfoutput=1

\documentclass[11pt]{article}

\usepackage[preprint]{coling}

\usepackage{times}
\usepackage{latexsym}

\usepackage[T1]{fontenc}

\usepackage[utf8]{inputenc}

\usepackage{microtype}

\usepackage{inconsolata}

\usepackage{graphicx}
\usepackage[table,xcdraw]{xcolor}
\usepackage{multirow}
\usepackage{scrextend}
\usepackage{xcolor}

\usepackage{arydshln}

\usepackage{amsmath}

\usepackage{amsfonts}

\title{A Scalable Framework for Automated NER Annotation Correction in Low-Resource Languages}

\author{
  \textbf{Toqeer Ehsan\textsuperscript{1}} and
  \textbf{Thamar Solorio\textsuperscript{2}}
\\
\textsuperscript{1}Quantitative Science and Technology Studies (QSTS),\\VTT Technical Research Centre of Finland Ltd., Espoo, Finland
\\
\textsuperscript{2}Mohamed bin Zayed University of Artificial Intelligence (MBZUAI),\\Masdar City, Abu Dhabi, United Arab Emirates
\\
\small{
\textbf{Correspondence:} \href{mailto:toqeer.ehsan@vtt.fi}{toqeer.ehsan@vtt.fi}
}
}

\begin{document}
\maketitle
\begin{abstract}
Poor quality or noisy annotations in Named Entity Recognition (NER), as in any other NLP task, make it challenging to achieve state-of-the-art performance. In this paper, we present a multi-step framework to enhance the annotation quality of NER datasets by employing automated techniques. We propose a frequency-based iterative approach that leverages self-training and a dual-threshold mechanism to enhance inference confidence. Experimental evaluations on different NER datasets demonstrate significant improvements in NER performance with respect to the original datasets. This work further explores the potential of generative Large Language Models (LLMs) to perform NER for low-resource languages.
\end{abstract}

\section{Introduction}
\label{sec:1}
With advancements in generative LLMs, multilingual models are now capable of performing several NLP tasks including NER. However, their performance on existing NER benchmarks, such as CoNLL-03 \cite{sang2003introduction} and OntoNotes 5.0 \cite{weischedel2013ontonotes}, remains moderate compared to supervised learning approaches \cite{chen2023robust, guo2024diluie, ye2023comprehensive}. 
The performance of supervised NER approaches is heavily dependent on the quality of the datasets, 
underscoring the need for well-annotated and improved datasets \cite{mayhew2023universal}.

\begin{table}[ht]
\small
\centering
\setlength\tabcolsep{3.5pt}
\renewcommand{\arraystretch}{1.3} 
\rowcolors{2}{gray!15}{white} 
\begin{tabular}{lccc}
\rowcolor{gray!30} 
\textbf{NE Type} & \textbf{Original NEs} & \textbf{Updated NEs} & \textbf{Increase (\%)} \\
\hline
\rowcolor{gray!20} 
\multicolumn{4}{c}{\textbf{MK-PUCIT}} \\
\textbf{PER} & 10,486 & 11,965 & 12.40\% \\
\textbf{LOC} & 19,868 & 23,880 & 16.80\% \\
\textbf{ORG} & 5,277 & 8,665 & 39.10\% \\
\textbf{Total NEs} & 35,631 & 44,510 & 19.90\% \\
\textbf{\# Sents.} & 24,080 & -- & --\\
\hline
\rowcolor{gray!20} 
\multicolumn{4}{c}{\textbf{Shahmukhi}} \\
\textbf{PER} & 4,609 & 4,732 & 2.60\% \\
\textbf{LOC} & 1,852 & 2,160 & 14.30\% \\
\textbf{ORG} & 526 & 644 & 18.30\% \\
\textbf{Total NEs} & 6,987 & 7,536 & 7.30\% \\
\textbf{\# Sents.} & 13,412 & -- & -- \\
\hline
\rowcolor{gray!20} 
\multicolumn{4}{c}{\textbf{SiNER}} \\
\textbf{PER} & 12,720 & 13,131 & 3.13\% \\
\textbf{LOC} & 14,136 & 15,029 & 5.94\% \\
\textbf{ORG} & 1,342 & 3,380 & 60.29\% \\
\textbf{Total NEs} & 28,198 & 31,540 & 10.60\% \\
\textbf{\# Sents.} & 31,612 & -- & -- \\
\hline
\end{tabular}
\caption{\normalsize Estimated number of named entities (PER, LOC, ORG) in the original training sets, the counts after filling missing annotations, and the percentage increase in named entities after insertion.}
\label{Tab:1}
\end{table}

Table~\ref{Tab:1} presents the statistics of the training sets of three NER datasets; MK-PUCIT \cite{kanwal2019urdu}, Shahmukhi \cite{ahmad2020named, tehseen2023shahmukhi}, and SiNER \cite{ali2020siner} for Urdu, Shahmukhi (Western Punjabi), and Sindhi languages.
Despite the large number of sentences, these datasets exhibit a significant amount of missing annotations, leading to reduced model performance on the NER task.
To gain a preliminary understanding of the extent of missing annotations in the datasets, we conducted an initial analysis by filling in potential missing entity mentions (see the estimates in Table~\ref{Tab:1}). This was achieved through context-free mapping of named entities extracted from the MK-PUCIT dataset onto all other training sets, including MK-PUCIT itself. This approach provided an estimate of the annotation gaps and highlighted the scale of the problem.
Manual revision and re-annotation, which are labor-intensive tasks requiring expertise and resources, are not feasible for datasets with such a large number of samples. These challenges highlight the need for automated approaches that can detect and fill missing annotations, improving the overall quality and completeness of these datasets.

We propose a frequency-based iterative technique to correct missing annotation errors using a self-training mechanism.
To ensure reliable predictions, a dual confidence threshold is imposed on logits and self-attention scores to reduce the risk of propagating erroneous annotation. Multiple annotation candidates are generated for each sentence using gold and pseudo-labels, and a multilingual NER model is employed as a validator to select the most plausible candidate based on the F\textsubscript{1} score. Before employing the self-training model, all three datasets were enhanced automatically to ensure correct word segmentation and named entity augmentation for frequent entity mentions.

The performance of NER on the corrected datasets is evaluated against baseline (original) datasets by fine-tuning the multilingual XLM-RoBERTa-large \cite{DBLP:journals/corr/abs-1911-02116, liu2019roberta} model.
The corrected datasets demonstrate performance improvements of 3.96, 1.4 and 1.44 micro F\textsubscript{1} points for MK-PUCIT, Shahmukhi and SiNER, respectively. Our approach effectively integrates the robustness of multilingual NER models with iterative self-training to enhance annotation quality of low-resource NER datasets. The main contributions of this work are as follows:

\begin{itemize}
\item We propose a frequency-based iterative technique that effectively corrects missing annotation errors in NER datasets
\item We demonstrate the capability of causal LLMs to improve the quality of non-English datasets.
\item We provide insights into the potential of causal LLMs for NER in low-resource languages by employing an in-line NER labeling method.
\item We prepare accurate validation and test sets for all three datasets, ensuring unbiased and reliable performance assessments.
\end{itemize}

\begin{figure*}[ht]
\centering
  \includegraphics[width=\linewidth]{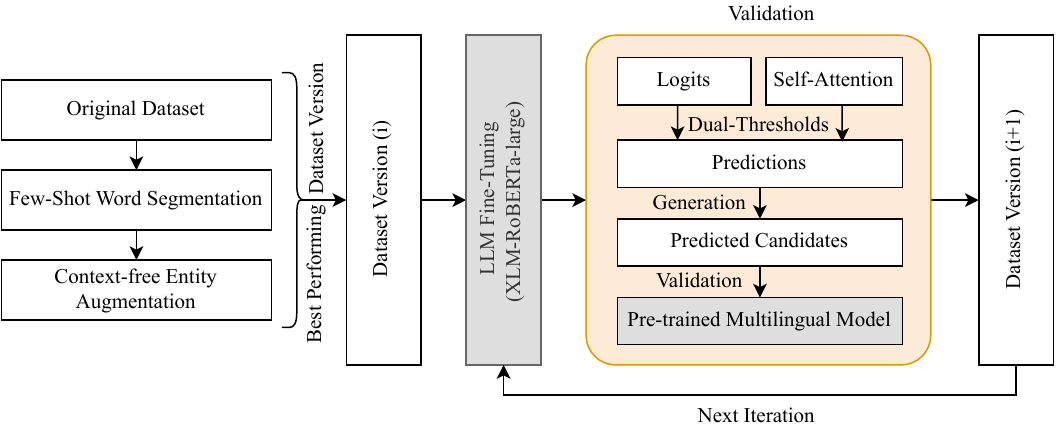}
  \caption{{Framework for NER Dataset Correction: An iterative process leveraging word segmentation correction, context-free entity augmentation, and iterative self-training with dual-threshold validation to enhance the annotation quality.}}
  \label{fig:architecture}
\end{figure*}

\section{Related Work}
\label{sec:2} 
\subsection{High-Resource NER Correction}
\label{sec:2-1}
Significant efforts have been made to develop advanced NER systems, while comparatively less attention has been paid to improving the quality of the annotation of the available datasets on which these models are based \cite{bernier2024annotation}. The quality of annotated data has a strong impact on NER performance. Recent research has focused on cleaning and correcting high-resource corpora such as OntoNotes 5.0 \cite{weischedel2013ontonotes}, ANERcorp \cite{benajiba2007anersys}, and CoNLL-03 \cite{sang2003introduction}. \citet{bernier2024annotation} addressed annotation errors in OntoNotes 5.0, a major English NER dataset, by identifying errors and applying automated correction rules. Their re-annotated dataset improved overall F-scores but required substantial manual effort.
Similarly, \citet{al2024cleananercorp} introduced CLEANANERCorp, a refined version of ANERcorp for Arabic NER. Using CLEANLAB \cite{wang2022detecting}, they detected mislabeled entities and manually re-annotated them with refined guidelines.
\citet{rueda2024conll} introduced CoNLL\#, a revised version of the CoNLL-03 English test set to support fine-grained error analysis. In contrast, \citet{rucker2023cleanconll} released CleanCoNLL, a semi-automatic relabeling of CoNLL-03 with updated and more consistent NER labels across train, dev, and test, assisted by entity linking and automatic consistency checks, followed by iterative cross-checking with manual inspection and correction. These approaches demonstrate that manual, semi-automatic, consistency-checked corrections can improve datasets, but they remain applicable mainly to well‑resourced languages.
\textit{Limitations:} High-resource correction approaches generally assume availability of external resources like Wikipedia and expert annotators. Consequently, they are difficult to apply to low‑resource languages, where annotated data and supporting resources are scarce.

\subsection{LLM-Assisted Annotation}
\label{sec:2-2}
\citet{yao2024can} compared human and LLM-generated annotations on a Chinese address NER dataset, finding that LLMs performed comparable to humans for building-level entities but underperformed for other entities. 
\citet{naraki2024augmenting} combined manual annotation with ChatGPT few-shot prompting in subsets of CoNLL03 and WikiGold \cite{balasuriya2009named}, showing that careful prompt design can reduce human effort, but still requires manual verification.
\citet{tan2024large} surveyed LLM‑aided annotation and synthesis strategies, and \citet{zhang2023llmaaa} proposed treating LLMs as ``active annotators''. These studies highlight the promise of LLMs to accelerate annotation, but also underline the need for careful calibration and human oversight.
\textit{Limitations:} Existing LLM‑assisted methods require prompt engineering and manual annotation verification. Moreover, the computational cost and less exposure to low-resource regional languages limit their scalability.

\subsection{Low-Resource NER Correction}
\label{sec:2-3}
For low‑resource languages, most work focuses on cross‑lingual transfer or data augmentation. Cross‑lingual and multilingual transfer techniques exploit character‑level models or multilingual embeddings to improve low‑resource NER \cite{cotterell2024low, torres2024experimental}. Cross-lingual data augmentation helps to improve NER performance for low-resource languages \cite{ehsan-solorio-2025-enhancing}. \citet{liu-etal-2025-improving} performed low-resource relation extraction and introduced logical rules to guide LLMs in producing high-confidence pseudo-labels achieving improved results.
\textit{Limitations:} These methods improve the robustness of the model but do not fill the annotation gaps. Distant supervision yields noisy labels, synthetic data can be culturally implausible, and cross-lingual augmentation introduces new samples without resolving missing annotation.



\section{Multi-Step Framework}
\label{sec:3}
The languages selected in this work are topologically related and culturally similar. In terms of named entities, they share similar names, locations and organizations. Given these similarities, cross-lingual representation could be helpful in improving the performance of NER for the regional languages. However, all three datasets contain a substantial amount of missing annotations, making it challenging for supervised models to achieve state-of-the-art performance. To address this issue, we propose a dataset correction framework that incorporates word segmentation correction, context-free named entity augmentation, and a frequency-based iterative self-training mechanism shown in Figure~\ref{fig:architecture}. Following sections describe the steps of our proposed approach.

\subsection{Word Segmentation}
\label{sec:3-0}
Urdu, Shahmukhi, and Sindhi are written in Perso-Arabic script, which is a cursive script \cite{unicode_arabic}. A character may have one of four positions within a token; initial, middle, final, or independent \cite{zia2018urdu, 10769409}. Characters are further categorized into joiners and non-joiners. Joiner characters are joined with the preceding character, while non-joiners are not. Importantly, the space character is not mandatory; instead, it is used to keep the correct shape of tokens. 

\begin{figure}[ht]
  \includegraphics[width=\linewidth]{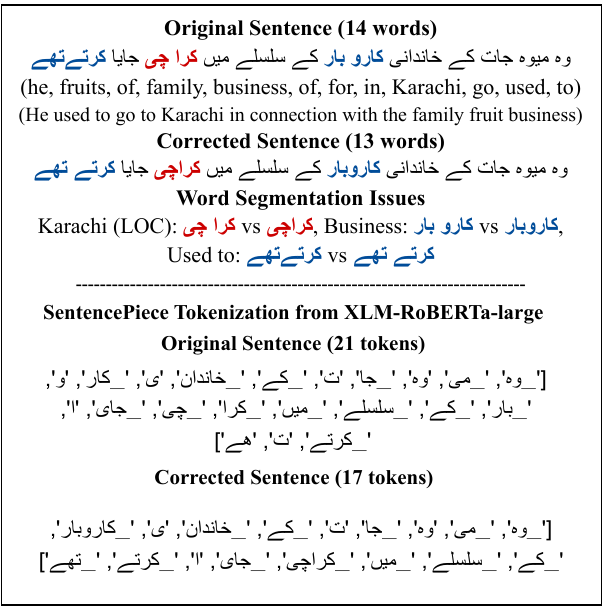}
  \caption{{An example from the MK-PUCIT dataset demonstrating the effect of inconsistent word segmentation on tokenization for transformer-based models.}}
  \label{fig:tokenization}
\end{figure}

Different writers, sources, and keyboards generate varying character sequences for the same content. LLM performance depends on tokenization, and inconsistencies can significantly impact model accuracy \cite{ovalle2024tokenization, kudo2018subword, schmidt2024tokenizatio}. All three datasets exhibit word segmentation errors, further complicating supervised NER model performance. Figure~\ref{fig:tokenization} illustrates an MK-PUCIT example with incorrect segmentation and its XLM-RoBERTa-large tokenization.

There are three space inconsistencies between original and corrected sentences in Figure~\ref{fig:tokenization}, that produce different tokenization from the SentencePiece tokenizer \cite{kudo2018sentencepiece}. The original sentence from the MK-PUICT produces 21 tokens, whereas the corrected sentence has 17 tokens. Along with others, there is a notable tokenization inconsistency in the named entity Karachi (LOC) due to incorrect word segmentation in the original dataset. This highlights that accurate word segmentation is crucial for consistent tokenization during the training and fine-tuning of LLMs.

To address word segmentation errors across all three datasets, we employed ChatGPT-4o (May 2024 version) using few-shot learning approach. We used the OpenAI API\footnote{platform.openai.com} to prompt the model to correct word segmentation while preserving the NER annotations. For this purpose, the CoNLL format was transformed into an in-line annotation format, inspired by \citet{paolini2021structured}. A transformed sequence is represented as:\\

\begin{addmargin}[1.5em]{1.5em}
\texttt{token-0 [token-1]NERTAG-1 token-2 [token-3 token-4]NERTAG-2 token-5.}\\
\end{addmargin}

The entity text span is enclosed in square brackets, followed by its corresponding NER tag (e.g., PER, LOC, or ORG). All three datasets were updated with correct word segmentation for further processing. The prompt used for in-context word segmentation is shown below.\\

\begin{addmargin}[1.0em]{0em}
{\textbf{Prompt:} ``You are an expert in identifying correct word boundaries in \{language name\}. Your task is to perform word segmentation by inserting missing blank spaces between words and by removing extra blank spaces. The text also contains named entities enclosed in square brackets followed by an entity label.\\
Instructions:\\
- Insert missing blank spaces between two or more words.\\
- Remove wrong blank spaces that separate characters of a single word.\\
- Perform word boundary correction within square brackets as well as outside of brackets.\\
- Keep the named entity annotation intact along with their labels.\\
- Punctuations should remain as space-separated tokens.\\
- Within square brackets, update boundaries only when the produced word is valid.\\
- Spellings and number of words must remain the same. Do not add any extra symbol, character, or punctuation.\\
Three examples are given for your reference.\\
Examples:\\
INPUT: \{\textit{sentence with few word seg. errors}\}\\
OUTPUT: \{\textit{sentence with correct word seg.}\}''}
\end{addmargin}

\subsection{Context-Free Entity Augmentation}
\label{sec:3-1}
The datasets are further enhanced by filling the missing annotations. For this purpose, entity lexicons are extracted for each entity type from each dataset, with minimum frequency threshold of three. These lexicons were then used to annotate unannotated sequences for each entity type without considering surrounding tokens/context. We compared the performance of models fine-tuned on the original, word segmented, and entity augmented datasets. The dataset version achieving the highest F\textsubscript{1} score on the validation set was selected for further processing.

\subsection{Iterative Self-Training}
\label{sec:3-2}
We propose a frequency-based iterative approach to address annotation errors in low-resource NER datasets. The approach focuses on refining the missing annotations by self-supervised learning inspired by \citet{el2022adasl}, enabling iterative improvements in annotations. The iterative approach operates on a cleaner version of the datasets produced from the previous steps. At each cycle, we fine-tune the model on the current labeled subset of the training split and then run inference on the full training split to propose missing entity labels.

To ensure reliable predictions, we employ a dual-thresholding mechanism that applies a confidence threshold on logits and self-attention scores. 
The refined versions of datasets are evaluated against validation sets by fine-tuning a multilingual XLM-RoBERTa-large model after each iteration. 

\subsection{Thresholds}
\label{sec:3-4}
The approach involves a probability threshold ($\tau_p$) and an attention-based dynamic threshold ($\tau_a$), which together reduce the possibility of over-propagating erroneous annotations.

\subsubsection{Probability Thresholding}
For a given token $i$, let $\mathbf{z}_i \in \mathbb{R}^T$ represent the logits for $T$ possible NER tags. The probability $p_{i,t}$ of the token $i$ being assigned tag $t$ is computed as:
\begin{equation}
p_{i,t} = \frac{\exp(z_{i,t})}{\sum_{t' \in \mathcal{T}} \exp(z_{i,t'})}, \label{eq:softmax}
\end{equation}

where $\mathcal{T}$ is the set of possible NER labels in the training set. The predicted tag for token $i$ is achieved by the highest probability as shown in Eq~\ref{eq:argmax}.
\begin{equation}
\hat{t}_i = \arg\max_{t \in \mathcal{T}} p_{i,t}. \label{eq:argmax}
\end{equation}
\begin{equation}
p_i = \max_{t \in \mathcal{T}} p_{i,t} \geq \tau_p, \label{eq:prob_threshold}
\end{equation}

For predictions with high confidence, only tokens that have a probability higher than or equal to the threshold $\tau_p$ are considered (Eq~\ref{eq:prob_threshold}).
Tokens that do not fulfill this criterion are assigned the non-entity tag (O). This step is helpful in reducing false positives predicted from low-confidence probabilities.

\subsubsection{Attention-Based Filtering}
To find the contextually significant predictions, we incorporate attention-based filtering by obtaining self-attention scores from the model's final layer. High self-attention scores indicate labels that have inherent meaning which are useful for NER, especially for noisy datasets as named entities often rely on their inherent meaning. This filtering approach selects semantically relevant NER labels by minimizing noise and incorrect annotations. Let $\mathbf{A} \in \mathbb{R}^{L \times L}$ denote the self-attention matrix for a sequence of length $L$, where $A_{i,j}$ represents the attention score from token $i$ to token $j$. The self-attention score for token $i$ is represented by $A_{i,i}$, corresponding to the diagonal elements of the attention matrix. We compute a dynamic threshold $\tau_a$ based on the top 10\% of self-attention scores for a sentence. The $\tau_a$ is defined as the 90th percentile of self-attention scores along the diagonal of $\mathbf{A}$.
\begin{equation}
\tau_a = \text{Percentile}_{90}(\{A_{j,j} \mid j = 1, \dots, L\}). \label{eq:attn_threshold}
\end{equation}

Tokens with self-attention scores greater than or equal to $\tau_a$ are considered for prediction shown as $A_{i,i} \geq \tau_a$. This ensures that only the top 10\% of tokens with the highest contextual relevance are considered for further predictions. Tokens that do not meet this threshold criterion are filtered out, retaining only predictions with significant contextual importance.

To enhance the reliability of the final predictions, we combine the probability and attention-based filtering mechanisms. The combined rule for assigning the final tag $\hat{t}_i^{\text{F}}$ to token $i$ is shown in Eq~\ref{eq:combined_filtering}.
\begin{equation}
\hat{t}_i^{\text{F}} =
\begin{cases}
\hat{t}_i, & \text{if } p_i \geq \tau_p \text{ and } A_{i,i} \geq \tau_a, \\
O, & \text{otherwise}.
\end{cases} \label{eq:combined_filtering}
\end{equation}


The proposed thresholding balances precision and recall in the iterative self-training approach. The threshold on the NER labels, ensures the statistical confidence by discarding low-confidence predictions. 

\begin{figure}[ht]
  \includegraphics[width=\linewidth]{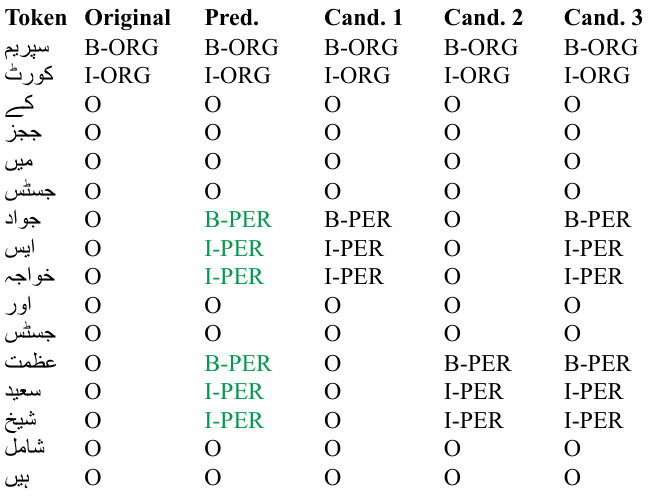}
  \caption{{An example of candidate generation from predicted pseudo labels. Translation: The judges of the Supreme Court include Justice \textbf{Jawad S. Khawaja} and Justice \textbf{Azmat Saeed Sheikh}.}}
  \label{fig:candidates}
\end{figure}

\subsubsection{Validation of Pseudo-Labels}
A multilingual XLM-RoBERTa-large NER model fine-tuned on all three datasets is employed to validate and refine the pseudo-label sequences predicted during the self-training process. For each predicted sentence, the pseudo-label sequence \( \hat{Y}_p \) is used to generate multiple candidate sequences \( \{C_1, C_2, \ldots, C_n\} \), along with the gold labels \( Y_{\text{orig}} \). 
These candidates are generated using a permutation-based approach if more than one named entity is identified, as shown in Figure~\ref{fig:candidates}. For example, if two named entities are introduced by the self-training iterative model, three candidates are generated: one with the first named entity, the second with the second named entity, and the third with both named entities.
Each candidate is evaluated against the predictions of the multilingual NER model, and the sequence with the highest \( F_1 \)-score is selected for integration into the dataset.
The model's raw predictions could be noisy; the candidate selection refines the results and filters out lower-confidence predictions. By using this validation, we effectively leverage the predictions while ensuring consistency and robustness in the final labels.
The candidate with the highest \( F_1 \)-score is selected as:
\begin{equation}
C_{\text{best}} = \arg\max_{C_i} F_1(C_i).
\label{eq:argmaxf1}
\end{equation}

The validation step ensures robust evaluation of pseudo-labels using the multilingual NER model. It incorporates accurate label sequences into the enhanced datasets that iteratively improve annotation quality.

\section{Datasets and Languages}
\label{sec:4}
Pakistani languages pose several challenges for the task of NER, such as absence of capitalization, contextual ambiguity, flexible word-order, and agglutinating nature \cite{khalid2023using, ehsan2021development, ahmed2024enriching}.
\textcolor{black}{Despite the larger sample sizes in MK-PUCIT, Shahmukhi, SiNER datasets, they face limited domain coverage, incomplete NER labels, low sentence-to-entity ratio, and noisy annotations, highlighting their low-resource status.} 

In this paper, we enhance three existing datasets, MK-PUCIT, Shahmukhi-NER and SiNER. 
For evaluation purposes, we manually reviewed NER annotations for the validation and test splits of all three datasets using the annotation guidelines provided by \citet{kanwal2019urdu}. We maintained the integrity of the validation and test sets by keeping them independent of the propagation process to ensure unbiased benchmarks for evaluation. The evaluation sets were manually curated, and therefore, not affected by any label propagation. 
We used 100 reference sentences from each evaluation set to compute the inter-annotator agreement (IAA) between an expert linguist and the annotator (first author).
The statistics of the evaluation sets are given in Table~\ref{tab:evalsets}. 
Appendix~\ref{app:a1} (Table~\ref{tab:eval-stats2}) reports and analyzes label-count changes by comparing the original vs. revised named-entity counts (PER/LOC/ORG) for both validation and test splits of all three datasets.
We used validation sets to evaluate the quality of each dataset after each processing step and final results are computed on test sets.

\begin{table}[ht]
\small
\centering
\setlength\tabcolsep{2pt}
\begin{tabular}{lcccccc}
\hline
\textbf{Eval. Sets} & \multicolumn{2}{c}{\textbf{MK-PUCIT}} & \multicolumn{2}{c}{\textbf{Shahmukhi}} & \multicolumn{2}{c}{\textbf{SiNER}} \\
\textbf{NE Type} & \textbf{Val} & \textbf{Test} & \textbf{Val} & \textbf{Test} & \textbf{Val} & \textbf{Test} \\
\hline
PER & 760 & 1,330 & 412 & 1,210 & 917 & 985 \\
LOC & 892 & 2558 & 146 & 432 & 884 & 1,106 \\
ORG & 190 & 808 & 54 & 177 & 211 & 185 \\
TOTAL & 1,842 & 4,696 & 612 & 1,819 & 2,012 & 2,276 \\
\#Sents. & 1,146 & 2,201 & 1,000 & 1,000 & 1,000 & 1,000 \\
\hline
IAA & 0.9713 & 0.9765 & 0.9604 & 0.9551 & 0.9440 & 0.9615 \\
\hline
\end{tabular}
\caption{{Statistics of manually revised evaluation sets for all three datasets and inter-annotator agreement using Cohen's Kappa scores on 100 reference sentences.}}
\label{tab:evalsets}
\end{table}

\paragraph{Urdu:}
Urdu is relatively resource-rich compared to other regional languages and is in the \textit{Vital} category \cite{eberhard2024ethnologue}. Several NER datasets are available for Urdu with different data annotations and sizes \cite{khana2016named, hussain2008resources, jahangir2012n, malik2017urdu}. However, we experimented with MK-PUCIT \cite{kanwal2019urdu}, which is annotated with coarse-grained named entities: person, location, and organization and is a larger dataset.

\paragraph{Shahmukhi:}
We experimented with the only available dataset for Shahmukhi, annotated with person, location, and organization labels \cite{ahmad2020named}. The quality of the dataset was previously enhanced by using the BIO annotation scheme \cite{tehseen2023shahmukhi}, and we use it as the Shahmukhi benchmark in our experiments. 

\paragraph{Sindhi:}
The SiNER dataset is the first large NER dataset for Sindhi \cite{ali2020siner}. Although it contains additional entity types, we experimented with three coarse-grained entity categories to ensure compatibility with the other datasets (MK-PUCIT and Shahmukhi NER). 

\section{Experimental Setup}
\label{sec:5}
We conduct experiments that are designed to improve the annotation quality of NER datasets for low-resource languages, where generative LLMs often struggle due to limited language representation \cite{naguib2024few, villena2024llmner, lu2024large, chen2023robust}. This research addresses two key questions; 1) How can supervised models be used to improve the annotation errors of noisy datasets? 2) How effective are generative LLMs to perform NER for low-resource languages? 
We hypothesize that supervised self-training mechanism can enhance NER annotations without requiring manual re-annotation or evaluation. Furthermore, if generative LLMs can perform NER comparably well, they can be employed as alternative annotation tools.

\subsection{Supervised NER Model}
For our experiments, we employed the pre-trained multilingual model, XLM-RoBERTa-large \cite{DBLP:journals/corr/abs-1911-02116}, which is based on RoBERTa's architecture \cite{liu2019roberta}. 
XLM-RoBERTa-large is a transformer-based model that supports 100 languages, including Urdu, Sindhi, and Shahmukhi script. It is pre-trained on large multilingual text corpora using Masked Language Modeling (MLM) objective. 

To fine-tune the model for NER, we added a token classification on top of the final transformer layer, which receives the hidden states from the last layer of the model, and computes the multi-class probability distribution over the entity classes for each token. Hyperparameters are described in \ref{app:a2}. Additionally, thresholding on label probabilities and self-attention scores is performed on the outputs from the final hidden layer, as described in Section~\ref{sec:3-2}.


\subsection{Zero/Few-Shot NER}
While the primary focus of this paper is on the enhancement of NER datasets utilizing supervised models, we also explore few-shot learning as an alternative approach. For this purpose, we transform the CoNLL format into an in-line format as described in Section~\ref{sec:3-0}.
We employed the ChatGPT-4o model, a state-of-the-art multilingual generative model, using the following prompt for few-shot learning.\\

\begin{addmargin}[1.0em]{0em}
{\textbf{Prompt:} ``You are an expert in identifying and annotating Named Entities in \{language name\}. Your task is to annotate three named entities, PERSON (PER), LOCATION (LOC), and ORGANIZATION (ORG) as per the following guidelines:\\
PER: Names of persons without job titles and designations such as, \{title examples\}. Include titles that represent caste, tribe or religious titles such as, \{title examples\}.\\
LOC: Names of places, cities, towns, countries etc.\\
ORG: Names of organizations, public or private.\\
Instructions:\\
- Number of words must remain the same in the INPUT and OUTPUT sentences.\\
- Respond with only annotated output without any additional description.\\
Annotate sentences according to the format in the following five examples.\\
Examples:\\
INPUT: \{Unannotated sentence\}\\ 
OUTPUT: \{Annotated sentence\}'' }
\end{addmargin}

\subsection{Iterative Self-training Protocol}
Our iterative model is run for 10 cycles, and macro-F\textsubscript{1} is computed after each iteration on the validation set. The first iteration is fine-tuned on training sentences containing named entities with a minimum frequency of three, which is gradually reduced to one for later iterations. In each cycle, we fine-tune on the current non-empty subset of the training split (sentences with at least one named entity) and then predict over the full training split to propose additional labels (no cross-fold partitioning); proposed labels are filtered by our thresholds (Section~\ref{sec:3-4}) before being merged into the next dataset version. We exclude fully empty sentences because they may contain unannotated entities that could introduce noise and degrade performance, and we select the final dataset version using the best validation macro-F\textsubscript{1} across cycles.

\section{Results and Discussion}
\label{sec:6}
We report results on the manually revised validation/test splits described in Section~\ref{sec:4} (Table~\ref{tab:evalsets}), first analyzing LLM-based zero/few-shot NER and then the supervised correction pipeline.

\subsection{Zero/Few-Shot NER Analysis}
Our NER experiments begin with in-context learning, employing zero-shot and few-shot learning using ChatGPT-4o. We evaluated generated outputs for both validation and test sets to demonstrate model's capability to perform NER and annotation correction. Table~\ref{tab:chatgptresults} presents results for all three datasets. 

\begin{table}[ht]
\centering
\small
\setlength\tabcolsep{4pt}
\begin{tabular}{lcccccc}
& \multicolumn{3}{c}{\textbf{Zero-shot NER}} &\multicolumn{3}{c}{\textbf{Few-shot NER}} \\
\hline
\textbf{NE Type} & \textbf{Pre.} & \textbf{Rec.} & \textbf{$F_1$}& \textbf{Pre.} & \textbf{Rec.} & \textbf{$F_1$} \\
\multicolumn{7}{c}{\textbf{Validation Sets}}\\
\hline
MK-PUCIT & 70.06 & 68.74 & 69.40 & 75.14 & 73.23 & 74.19 \\
Shahmukhi & 59.09 & 70.70 & 64.38 & 67.32 & 78.56 & 72.51 \\
SiNER & 62.76 & 73.88 & 67.87 & 67.22 & 79.86 & 73.54 \\
\multicolumn{7}{c}{\textbf{Test Sets}}\\
\hline
MK-PUCIT & 73.47 & 71.73 & 72.58 & 72.12 & 80.64 & 76.14 \\
Shahmukhi & 63.84 & 67.71 & 65.72 & 65.49 & 76.66 & 70.63\\
SiNER & 64.92 & 75.68 & 69.89 & 71.21 & 83.57 & 76.90\\
\hline
\end{tabular}
\caption{{Zero-shot and few-shot learning based micro-F\textsubscript{1} scores on validation and test sets for all three datasets using ChatGPT-4o.}}
\label{tab:chatgptresults}
\end{table}

\begin{table}[ht]
\centering
\small
\setlength\tabcolsep{10pt}
\begin{tabular}{llll}
\hline
\textbf{Datasets} & \textbf{Pre.} & \textbf{Rec.} & \textbf{$F_1$} \\
\multicolumn{4}{c}{\textbf{MK-PUCIT Validation}} \\
\hline
Original & 80.59 & 72.45 & 76.31 \\
Original+WS & 81.09 & 73.05 & 77.08 \\
Original+WS+AUG & 78.23 & 77.70 & \textbf{77.96} \\
\multicolumn{4}{c}{\textbf{Shahmukhi Validation}} \\
\hline
Original & 81.74 & 77.19 & 79.40 \\
Original+WS & 84.89 & 77.71 & \textbf{81.14} \\
Original+WS+AUG. & 77.36 & 79.36 & 78.34 \\
\multicolumn{4}{c}{\textbf{SiNER Validation}} \\
\hline
Original & 93.52 & 80.80 & 86.70 \\
Original+WS & 94.31 & 80.93 & 87.11 \\
Original+WS+AUG & 89.46 & 87.90 & \textbf{88.67} \\
\hline
\end{tabular}
\caption{{Micro-F\textsubscript{1} scores evaluated on the validation sets for each dataset using the original, word-segmented, and augmented versions.}}
\label{tab:validationresults}
\end{table}

\subsection{Automated Annotation Correction}
Table~\ref{tab:validationresults} presents baseline results obtained by fine-tuning the XLM-RoBERTa-large model on the original versions of all three datasets. While the performance of ChatGPT-4o is quite impressive on these low-resource languages, the supervised multilingual model still outperforms it, even when fine-tuned on the original datasets containing missing annotations.

The performance of the supervised model is further improved with word segmentation correction and entity augmentation. All three datasets demonstrate performance enhancement with correct word segmentation that leads to more accurate tokenization. 
Unannotated named entities were filled using frequent named entities extracted from all three datasets. The MK-PUCIT and SiNER demonstrate improvements with named entity augmentation, whereas the Shahmukhi dataset shows a decrease.
\begin{table}[ht]
\small
\centering
\begin{tabular}{lccccc}
\multicolumn{6}{c}{\textbf{MK-PUCIT}}\\
\textbf{Iter.} & \textbf{Pre.} & \textbf{Rec.} & \textbf{F-Score} & \textbf{\#Sents.} & \textbf{\#NEs} \\
\hline
0 & 76.50 & 39.96 & 49.66 & 12,412 & 25,479 \\
1 & 75.05 & 49.46 & 56.66 & 14,301 & 30,740 \\
2 & 78.18 & 70.96 & 74.28 & 18,629 & 44,621 \\
3 & 78.47 & 70.91 & 74.00 & 18,658 & 44,655 \\
4 & 78.17 & 77.21 & \textbf{77.68} & 18,677 & 44,677 \\
5 & 75.63 & 78.31 & 76.84 & 18,701 & 44,712 \\
6 & 78.08 & 75.90 & 76.94 & 18,736 & 44,749 \\
7 & 76.19 & 71.40 & 73.34 & 18,745 & 44,759 \\
8 & 78.79 & 67.79 & 72.45 & 18,750 & 44,768 \\
9 & 80.80 & 69.81 & 74.64 & 18,763 & 44,792 \\
\hline
\multicolumn{6}{c}{\textbf{Shahmukhi}}\\
\textbf{Iter.} & \textbf{Pre.} & \textbf{Rec.} & \textbf{F-Score} & \textbf{\#Sents.} & \textbf{\#NEs} \\
\hline
0 & 87.66 & 65.05 & 73.93 & 2,502  & 3,403  \\
1 & 81.66 & 71.97 & 76.47 & 3,149  & 4,505  \\
2 & 84.62 & 77.79 & 81.03 & 4,439  & 7,105  \\
3 & 81.28 & 78.28 & 79.50 & 4,471  & 7,137  \\
4 & 79.06 & 76.26 & 77.49 & 4,476  & 7,149  \\
5 & 82.62 & 77.02 & 79.44 & 4,782  & 7,449  \\
6 & 81.44 & 76.86 & 78.73 & 5,167  & 7,834  \\
7 & 76.99 & 80.18 & 78.40 & 5,189  & 7,856  \\
8 & 84.33 & 79.07 & \textbf{81.52} & 6,010  & 8,679  \\
9 & 83.45 & 73.10 & 77.52 & 7,009  & 9,680  \\
\hline
\multicolumn{6}{c}{\textbf{SiNER}}\\
\textbf{Iter.} & \textbf{Pre.} & \textbf{Rec.} & \textbf{F-Score} & \textbf{\#Sents.} & \textbf{\#NEs} \\
\hline
0 & 77.94 & 70.71 & 74.11 & 9,838  & 17,275 \\
1 & 81.33 & 69.65 & 75.01 & 11,149 & 20,280 \\
2 & 80.55 & 79.50 & 80.02 & 13,913 & 28,997 \\
3 & 79.29 & 80.21 & 79.72 & 14,023 & 29,085 \\
4 & 83.03 & 87.38 & \textbf{84.83} & 14,077 & 29,139 \\
5 & 82.76 & 81.14 & 81.81 & 14,242 & 29,305 \\
6 & 80.19 & 75.80 & 77.89 & 14,335 & 29,398 \\
7 & 81.40 & 76.87 & 79.03 & 14,372 & 29,435 \\
8 & 79.51 & 84.49 & 81.16 & 14,457 & 29,523 \\
9 & 83.12 & 82.87 & 82.96 & 14,516 & 29,582 \\
\hline
\end{tabular}
\caption{{Macro-F\textsubscript{1} scores for all three datasets from our iterative self-training alongside number of sentences and named entities after each iteration. Best F\textsubscript{1} scores are highlighted in bold.}}
\label{tab:iter_results}
\end{table}

Table~\ref{tab:iter_results} presents F\textsubscript{1} scores on the validation set along with the number of sentences and the count of named entities. 
The sentence count keeps increasing as new named entities are introduced after each iteration, and the training set includes only sentences that contain at least one named entity. Based on these results, we selected the best-performing version of each dataset.

\begin{table}[ht]
\centering
\small
\setlength\tabcolsep{2pt}
\begin{tabular}{llccccc}
\hline
\multicolumn{1}{l}{} & \multicolumn{2}{c}{\textbf{Original Dataset}} & \multicolumn{4}{c}{\textbf{Improved Dataset}} \\
\multicolumn{1}{l}{\textbf{Datasets}} & \textbf{Val-F\textsubscript{1}} & \textbf{Test-F\textsubscript{1}} & \textbf{Val-F\textsubscript{1}} & \textbf{$\Delta$} & \textbf{Test-F\textsubscript{1}} & \textbf{$\Delta$} \\
\hline
\multirow{1}{*}{MK-PUCIT} & 76.31 & 84.59 & 80.68 & +4.37 & 88.55 & +3.96 \\
\multirow{1}{*}{Shahmukhi} & 79.40 & 81.34 & 80.21 & +0.81 & 82.74 & +1.40 \\
\multirow{1}{*}{SiNER} & 86.70 & 87.92 & 89.14 & +2.44 & 89.36 & +1.44 \\
\hline
\end{tabular}
\caption{{Micro-F\textsubscript{1} scores of the best-performing datasets on the test sets alongside the original versions of each dataset.}}
\label{tab:finalresults}
\end{table}


MK-PUCIT achieves the best macro-F\textsubscript{1} score after the fifth iteration, Shahmukhi after the ninth iteration, and SiNER after the fifth iteration. The final results on the test sets are presented in Table~\ref{tab:finalresults}, using the dataset versions with highest scores from Table~\ref{tab:iter_results}, alongside a comparison with the micro-F\textsubscript{1} scores from the original datasets.  Table~\ref{tab:finalstats} presents the statistics of named entities for different types in the original and updated datasets. 
All entity types show increased counts in the updated datasets, suggesting that our iterative self-training approach effectively recovers missing named entities. However, counts alone do not confirm the correctness of each added label; improved F1 scores on clean test sets provide supporting evidence of better annotation quality. Additional qualitative examples of annotation corrections and their iterative updates are provided in Appendix~\ref{app:b}.

\begin{table}[ht]
\centering
\small
\setlength\tabcolsep{4pt}
\begin{tabular}{llcccc}
\hline
\multicolumn{1}{l}{\textbf{Dataset/Type}} & \textbf{Version} & \textbf{PER} & \textbf{LOC} & \textbf{ORG} & \textbf{Total} \\
\hline
\multirow{2}{*}{\textbf{MK-PUCIT}} & Original & 10,486 & 19,868 & 5,277 & 35,631 \\
 & Updated & 12,216 & 23,592 & 8,869 & 44,677 \\
 \hline
\multirow{2}{*}{\textbf{Shahmukhi}} & Original & 4,609 & 1,852 & 526 & 6,987 \\
 & Updated & 6,226 & 1,893 & 560 & 8,679 \\
 \hline
\multirow{2}{*}{\textbf{SiNER}} & Original & 12,720 & 14,136 & 1,342 & 28,198 \\
 & Updated & 12,753 & 13,958 & 2,428 & 29,139 \\
 \hline
\end{tabular}
\caption{{Statistics of type-wise named entities before and after entity propagation process.}}
\label{tab:finalstats}
\end{table}


Experimental results strongly support the first part of our hypothesis. Supervised self-training approach, combined with validation, achieved significant improvements in annotation and NER performance without any manual effort. These findings highlight the effectiveness of our approach which is suitable for the enhancement of low-resource sequence labeling datasets. However, the results from generative LLMs, such as ChatGPT-4o are impressive, but reveal limitations of its capabilities to perform NER for low-resource languages.

\section{Conclusion}
\label{sec:7}
Our proposed iterative self-training approach with dual-threshold mechanism and multilingual validation, demonstrates significant improvements in annotation quality and NER performance for three datasets, MK-PUCIT, Shahmukhi, and SiNER. Our approach effectively addresses the challenge of missing annotations for low-resource languages. Experimental results highlight robustness of the approach with consistent F\textsubscript{1} improvements. The methodology eliminates the need of labor-intensive manual re-annotation that makes it suitable for large-scale datasets in under-resourced languages. The work not only provides a scalable solution for noisy datasets, but also highlights the importance of high-quality datasets for robust NER systems.\\ 

\section*{Limitations}
\label{sec:7}
The quality of automated annotation depends on the performance of the self-trained model and the validation process. While the iterative approach demonstrates progressive refinement of empty annotations, errors in the initial pseudo-labels can propagate through iterations, leading to the reinforcement of incorrect annotations. Additionally, the approach is totally automated without any intervention from human-annotators. Inherently, it lacks the contextual understanding that human annotators can provide. Despite significant improvements, enhanced datasets may still contain some annotation inconsistencies.

\section*{Acknowledgments}
We acknowledge the use of ChatGPT (OpenAI) for writing assistance, grammar polishing, and text improvement.

\section*{Data Availability}
The updated versions of all three datasets after each iteration are available at: \url{https://github.com/toqeerehsan/low-res_ne_correction}.

\bibliography{bibliography}
\newpage

\appendix

\section{Appendix}
\label{app:a}

\subsection{Hyperparameters}
\label{app:a2}
In the self-training process, the learning rate of 2e-5 was used along with the AdamW optimizer. The batch size was set to 8, which helped to maintain memory and training efficiency. The models were fine-tuned by using a varying number of epochs for low-resource datasets depending on the training samples. Early stopping was further implemented based on the micro F\textsubscript{1} score on the validation set. The maximum sequence length was set to 100 tokens. These hyperparameters ensured optimal performance of the models.

\subsection{Attention Threshold Sensitivity Analysis}
\label{app:a3}

Table \ref{tab:ablation} reports the number of named entities extracted at various attention thresholds (percentiles 0.75–0.95). We conducted this analysis on a subset of the Shahmukhi dataset, trained for five epochs with the same settings as the first iteration (see Table \ref{tab:iter_results}). Because all three training sets contain many missing annotations, using a low attention threshold admits too many speculative candidates, which degrades the quality of the corrected dataset in later iterations. For this reason, we selected a 0.90 percentile threshold to retain only high-confidence named entities, and the results show that the threshold was helpful in improving the quality of the datasets.

\begin{table}[h]
\centering
\small
\begin{tabular}{lllll}
\hline
\textbf{Attention Threshold} & \textbf{PER} & \textbf{LOC} & \textbf{ORG} & \textbf{Total} \\
\hline
Percentile = 0.95  & 250 & 103 & 6 & 359\\
Percentile = 0.90  & 293 & 121 & 10 & 424\\
Percentile = 0.85  & 306 & 127 & 11 & 444\\
Percentile = 0.80  & 308 & 129 & 11 & 448\\
Percentile = 0.75  & 308 & 132 & 11 & 451\\
\hline
\end{tabular}
\caption{Number of detected named entities against different attention threshold percentiles.}
\label{tab:ablation}
\end{table}

\subsection{Correction Statistics for Evaluation Sets}
\label{app:a1}
Table~\ref{tab:eval-stats2} shows statistics of corrected/filled named entities in the validation and test sets for all three datasets.
\begin{table*}[ht]
\centering
\begin{tabular}{llllll}
\hline
& & \multicolumn{2}{c}{\textbf{Validation set}} &  \multicolumn{2}{c}{\textbf{Test Set}}\\
\multicolumn{1}{l}{\textbf{Dataset Name}} & \textbf{NE Type} & \textbf{Original} & \textbf{Revised} & \textbf{Original} & \textbf{Revised} \\
\hline
\multirow{4}{*}{MK-PUCIT} & PER & 657 & 760 & 1,275 & 1,330\\
 & LOC & 863 & 892  & 2,560 & 2,558 \\
 & ORG & 187 & 190  & 853 & 808  \\
 & Total & 1,707 & 1,842  & 4,688 & 4,696 \\
\hline
\multirow{4}{*}{Shahmukhi} & PER & 374 & 412 & 1229 & 1,210 \\
 & LOC & 128 & 146  & 428 & 432 \\
 & ORG & 49 & 54 &  128 & 177 \\
 & Total & 551 & 612  & 1,785 & 1,819 \\
\hline
\multirow{4}{*}{SiNER} & PER & 809 & 917 & 1101 & 985 \\
 & LOC & 908 & 884 & 1071 & 1,106 \\
 & ORG & 249 & 211 & 140 & 185 \\
 & Total & 1,966 & 2,012  & 2,312 & 2,276 \\
\hline
\end{tabular}
\caption{Statistics of named entities from validation and test sets after manual correction.}
\label{tab:eval-stats2}
\end{table*}

\section{Illustrative Annotation Corrections}
\label{app:b}
\begin{figure*}[ht]
\centering
  \fcolorbox{gray}{white}{\includegraphics[width=1.0\linewidth]{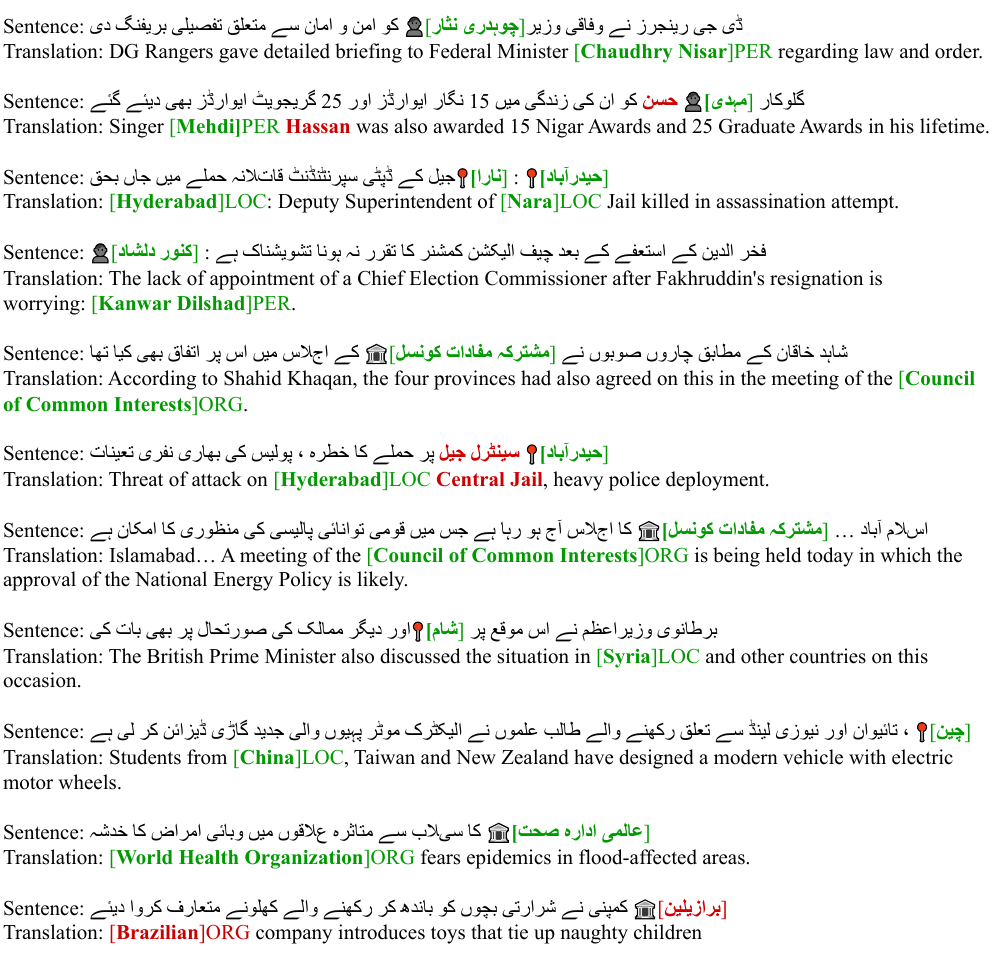}}
  \caption{Randomly selected sample sentences showing the annotations before and after label correction process. Corrected annotations are in green color and incorrect labels are in red color.}
  \label{fig:updated_examples}
\end{figure*}

\begin{figure*}[ht]
\centering
  \fcolorbox{gray}{white}{\includegraphics[width=1.0\linewidth]{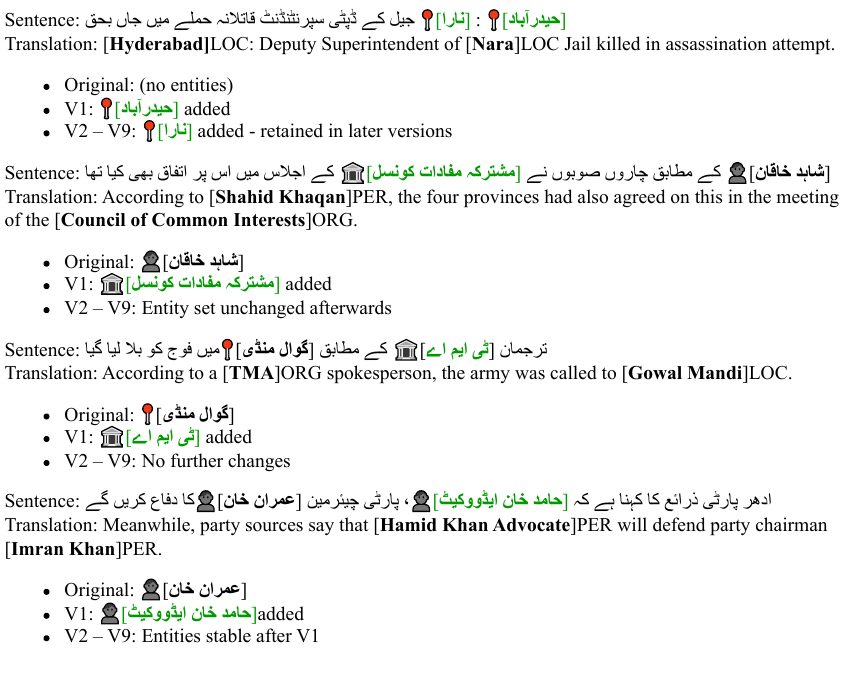}}
  \caption{Randomly selected sample sentences showing the annotations from different annotation versions. Correctly filled annotations are shown in green color.}
  \label{fig:versions_examples}
\end{figure*}

Figure~\ref{fig:updated_examples} presents annotation comparisons for some sample sentences before and after automatic label correction using the self-training label propagation method. Correctly identified labels are shown in green and incorrect labels in red. These examples were randomly selected from the corrected MK-PUCIT dataset. Although the dataset may still contain some incorrect annotations after automatic label correction and propagation, the overall quality of the dataset has improved compared to its original noisy annotations. Figure~\ref{fig:versions_examples} further shows sample sentences with different annotations from all ten versions of the MK-PUCIT dataset. The correctly filled empty annotations are colored green.  

\end{document}